\documentclass{article}

     \usepackage[final]{neurips_2019}

\usepackage[utf8]{inputenc} 
\usepackage[T1]{fontenc}    
\usepackage{hyperref}       
\usepackage{url}            
\usepackage{booktabs}       
\usepackage{amsfonts}       
\usepackage{nicefrac}       
\usepackage{microtype}      
\usepackage{times}
\usepackage{soul}
\usepackage{url}
\usepackage[utf8]{inputenc}
\usepackage[small]{caption}
\usepackage{graphicx}
\usepackage{amsmath}
\usepackage{booktabs}
\usepackage{algorithm}
\usepackage{algorithmic}
\usepackage{subfig}
\title{Layerwise Knowledge Extraction from Deep Convolutional Networks}

\author{Simon Odense \\
Department of Computer Science\\
City, University of London\\
London, EC1V 0HB, UK \\
\texttt{\{simon.odense\}@city.ac.uk} \\
\And
 Artur d'Avila Garcez \\
Department of Computer Science \\
City, University of London \\
London, EC1V 0HB, UK \\
\texttt{\{a.garcez\}@city.ac.uk} \\
}

\begin{document}

\maketitle
\begin{abstract}
Knowledge extraction is used to convert neural networks into symbolic descriptions with the objective of producing more comprehensible learning models. The central challenge is to find an explanation which is more comprehensible than the original model while still representing that model faithfully. The distributed nature of deep networks has led many to believe that the hidden features of a neural network cannot be explained by logical descriptions simple enough to be comprehensible. In this paper, we propose a novel layerwise knowledge extraction method using \textit{M-of-N} rules which seeks to obtain the best trade-off between the complexity and accuracy of rules describing the hidden features of a deep network. We show empirically that this approach produces rules close to an optimal complexity-error tradeoff. We apply this method to a variety of deep networks and find that in the internal layers we often cannot find rules with a satisfactory complexity and accuracy, suggesting that rule extraction as a general purpose method for explaining the internal logic of a neural network may be impossible. However, we also find that the softmax layer in Convolutional Neural Networks and Autoencoders using either \emph{tanh} or \emph{relu} activation functions is highly explainable by rule extraction, with compact rules consisting of as little as 3 units out of 128 often reaching over $99\%$ accuracy. This shows that rule extraction can be a useful component for explaining parts (or modules) of a deep neural network.
\end{abstract}

\section{Introduction}

Recently there has been an increase in interest in explainable Artificial Intelligence (AI). Although in the past decade there have been major advances in the performance of neural network models, these models tend not to be explainable \cite{NIPS2017}. In large part, this is due to the use of very large networks, specifically deep networks, which can contain thousands or even millions of hidden neurons. In contrast with symbolic AI, in which specific features are often hand picked for a problem, or symbolic Machine Learning (ML), which takes a localist approach \cite{Markov}, the hidden neurons in a deep neural network do not necessarily correlate with obviously identifiable features of the data that a human would recognise. 

Knowledge extraction seeks to increase the explainability of neural networks by attempting to uncover the knowledge that a neural network has learned implicitly in its weights. One way of doing this is to translate trained neural networks into a set of symbolic rules or decision trees similar to the ones found in symbolic AI, ML and logic programming \cite{textbook}. Over the years, many rule extraction techniques have been developed \cite{Towell1993}\cite{id2-of-3} \cite{Craven:1996:ECM:924305} \cite{Trann:2016} \cite{arturruleextraction} but none have been able to completely solve the black box problem for neural networks. The main barrier to comprehensible rule extraction is the complexity of the extracted rules. Even if it is possible to find a symbolic system which exactly describes a neural network, it may contain too many rules to be understandable.

Perhaps the main reason this has proved to be a difficult problem is that reasoning in neural networks takes place in a \textit{distributed} fashion \cite{DeepLearning:2016}. It has been argued that one of the fundamental properties of neural networks is that any abstract concepts it uses are represented in a distributed way, that is as patterns of activations across many hidden neurons rather than with a single hidden neuron  \cite{proper_connectionism}. 

The distributed nature of neural networks has led many to conclude that attempting to explain the hidden neurons of large neural networks using symbolic knowledge extraction is a dead end \cite{SoftTree:2017}. Instead, alternative approaches to explanation have grown in popularity (see \cite{Survey} for a survey). Such approaches are so varied that four distinct explainability problems have been identified: \emph{global explanations}, which attempt to give an explanation of a black box, \emph{local explanations}, which attempt to give an explanation for a particular output of a black box, \emph{visualization}, which gives a visual explanation of a latent feature or output, and \emph{transparent box design}, which seeks to create new models which have some inherent explainability. 

Recent trends have favoured \textit{model-agnostic} methods which opt to use the input-output relationship of a model to generate an explanation rather than assigning any meaning to hidden variables. From the point of view of transparency this may be adequate, but understanding the exact reasoning that a neural network uses with respect to its representation could shine new light into the kinds of knowledge that a deep neural network learns and how it uses that knowledge \cite{Garcez:2008:NCR:1478768}. This has the potential to accelerate the development of more robust models by illuminating any deficiencies that exist in current models and their learning algorithms.

In this paper, we develop a rule extraction method that can control for the complexity of a rule via the scaling of an objective function. We do this by performing a parallel search through the space of  \textit{M-of-N} rules \cite{Towell1993} and measuring the error and complexity of each rule. By restricting our search space and using parallel techniques we are able to apply our algorithm to much larger networks than more exhaustive search techniques. We evaluate our algorithm against an optimal search technique (CORELS \cite{CORELS}) on a series of small networks before applying it to the layers of deep convolutional networks. By selecting various error/complexity trade-offs, we are able to map out a rule extraction landscape which shows the relationship between how complex the extracted rules are allowed to be and how accurately they capture the behaviour of a network. We find that the relative explainability between layers differs greatly and that changes to the network such as activation function can affect whether or not rule extraction will be useful in certain layers.

In Section $2$, we provide an overview of previous algorithms used for knowledge extraction. In Section $3$, we give definitions of accuracy and complexity for \textit{M-of-N} rules and present the extraction algorithm. In Section $4$, experimental results are reported and discussed. Section $5$ concludes and discusses directions for future work.

\section{Background and Related Work}
Approaches to rule extraction can, in general, be identified as \textit{decompositional}, in which the parameters of the network are used to generate rules, \textit{pedagogical}, in which the behaviour of the network is used to generate rules, or \textit{eclectic} which are techniques with both decompositional and pedagogical components \cite{taxonomy}.
One of the first attempts at knowledge extraction used a decompositional approach applied to feedforward networks, in particular the Knowledge-based Artificial Neural Networks (KBANN) \cite{KBANN}. This algorithm used the weights of a hidden variable to extract symbolic rules of the form \emph{IF M out of a set of N neurons (or concepts) are activated (or hold) THEN a given neuron (concept) is activated (holds)}, called \textit{M-of-N} rules \cite{Towell1993}. This was followed by more sophisticated algorithms which generate binary trees in which each node is an \textit{M-of-N} rule \cite{id2-of-3} \cite{Craven:1996:ECM:924305} (Notice that these binary trees can be reduced to IF-THEN propositional logic sentences as before). These more recent algorithms are pedagogical in that they select an \textit{M-of-N} rule using the input units as the concepts (called \emph{literals} in logic), based on the maximum information gain with respect to the output. Other algorithms extract rules in the form of decision sets which are another rule based structure equivalent to decision trees. Two level decision sets have been used to generate both local explanations \cite{IDS} and global explanations \cite{DBLP:BDL}\cite{CORELS} but have only been done in a model-agnostic way with no attempt to explain the internal variables of a model such as the hidden neurons in a deep network.

Other methods abandon the knowledge extraction paradigm and opt for alternative techniques. In the context of computer vision, the use of visual importance methods might be preferred \cite{LIME}\cite{importance}. Another approach is to design models which are explainable by design \cite{VisualRecurrent} \cite{SoftTree:2017} \cite{Anchors} \cite{DBLP:journals/corr/CourbariauxB16}. In the last example, we note the similarity of the restricted model to \textit{M-of-N} rules, each hidden neuron in this case can be thought of as an \textit{M-of-N} rule.

Most decompositional rule extraction has been applied only to shallow networks. The multiple hidden layers in a deep network mean that in order to explain an arbitrary hidden feature in terms of the input, a decompositional technique has to produce a hierarchy of rules (see \cite{Trann:2016} for an example of hierarchical rule extraction). With many hidden layers, the extracted rules can quickly grow far too complex for a human to understand, unless each constituent of the rule hierarchy is exceedingly simple. Thus, the use of decompositional techniques to explain the features of a deep network end-to-end seems impractical, as argued in \cite{SoftTree:2017}. Nevertheless, experiments reported in this paper show that some layers of a deep network are associated with highly explainable rules opening up the possibility of rule extraction being used as a component in a modular explanation of network models.

\section{Layerwise Knowledge Extraction}
\subsection{\textit{M-of-N} Rules for Knowledge Representation:} 
In logic programming, a logical rule is an implication of the form $A \leftarrow B$, called $A$ \emph{if} $B$. The literal $A$ is called the \textit{head} of the rule and $B$ stands for a conjunction of literals, $B_1 \wedge B_2 \wedge ... \wedge B_n$ called the \textit{body} of the rule. Disjunctions in the body can be modelled simply as multiple rules having the same head. Most logic programs adopt a \textit{negation by failure} approach whereby $A$ is $true$ if and only if $B$ is $true$ \cite{LogicProgramming}. When using rules to explain a neural network, the literals will refer to the states of neurons. For example, if a neuron $x$ takes binary values \{0,1\} then we define the literal $X$ by $X=True$ if $x=1$, and $X=False$ if $x=0$. For neurons with continuous activation values, we can define a literal by including a threshold $a$ such that $X=True$ if $x > a$, and $X=False$ otherwise. In other words, the literal $X$ is shorthand for the statement $x>a$.

In neural networks, a hidden neuron is usually poorly described by a single conjunctive rule since there are many different input configurations which will activate a neuron. Rather than simply adding a rule for each input pattern that activates a neuron (which essentially turns the network into a large lookup table), we look for \textit{M-of-N} rules which have been commonly used in rule extraction starting with \cite{Towell1993}. \textit{M-of-N} rules soften the conjunctive constraint on the body of logical rules by requiring only $M$ of the variables in the body to be true for some specific value of $M<N$ (notice that when $M=N$ we are left with a conjunction). For example, the rule $H \leftarrow 2-of-\{X_1,X_2,\neg X_3\}$ is equivalent to $H \leftarrow (X_1 \wedge X_2)$ or $(X_2 \wedge \neg X_3)$ or ($X_1 \wedge \neg X_3)$, where $\neg$ stands for negation by failure.

\textit{M-of-N} rules are an attractive candidate for rule extraction because they share a structural similarity with neural networks. Indeed every \textit{M-of-N} rule can be thought of as a simple perception with binary weights and a threshold $M$. \textit{M-of-N} rules have been used in the early days of knowledge extraction but have since been forgotten. This paper brings \textit{M-of-N} rules to the forefront of the debate on explainability again.

When networks have continuous activation values, in order to define the literals to use for rule extraction we must choose a splitting value $a$ for each neuron which will lead to a literal of the form $x>a$. In order to choose such values for continuous neurons we use \textit{information gain} \cite{InormationGainDecisionTree}\cite{informationgain}
Given a target neuron $h$ that we wish to explain, we generate a literal for the target neuron by selecting a split based on the information gain with respect to the output labels of the network. That is, given a set of test examples, choose the value of the target neuron $h$ which splits the examples in such a way as to result in the maximum decrease in entropy of the network outputs on the test examples.

The input literals are then generated from the inputs to the target neuron by choosing splits for each input which maximize the information gain with respect to the target literal generated in the previous step. In practice this means that each target literal in a layer will have its own set of input literals, each corresponding to the same set of input neurons but with different splits.

In the case that the layer is convolutional, each feature map corresponds to a group of neurons, each with a different input patch. Rather than test every single neuron in the feature map we only test the one whose optimal split has the maximum information gain with respect to the network output. This gives us a single rule for each feature map rather than a collection of rules. 

\begin{algorithm}
\caption{Search procedure for finding \textit{M-of-N} rules to explain a hidden feature $h$}
\begin{algorithmic}

\STATE Generate a split, $s$, for $h$ by choosing the value which maximizes the information gain with respect to the network output. Use this to define the literal $H$

\FOR{Each neuron $x$ which is an input of $h$,} \STATE{Generate a split for $x$ by choosing the value which maximizes the information gain with respect to $H$. Use this value to define the literal $X$ if the connection between $x$ and $h$ is positive, and use it to define $\neg X$ otherwise}
\ENDFOR
\STATE Order the input literals by the magnitude of their weights
\FOR{$N:1 \leq N \leq$ \textit{number of inputs}}\FOR{$M:1\leq M \leq N$}\STATE{Create an $M-of-N$ rule, $R$, whose body consists of the first $N$ literals. Then compute $L(R)$;}\ENDFOR
\ENDFOR
\STATE{Compute $L(R)$ for the trivial rules $0-of-\{\}$ and $1-of-\{\}$;}
\RETURN{rule with the lowest value of $L(R)$.}
\end{algorithmic}
\end{algorithm}
\subsection{Soundness and Complexity Trade-off} The two metrics we are concerned with in rule extraction are comprehensibility and accuracy. For a given rule we can define accuracy in terms of a \textit{soundness} measure. This is simply the expected difference between the predictions made by the rules and the network. More concretely given a neuron $h$ in a neural network with input neurons $x_i$, we can use the network to compute the state of $h$ from the state of the input neurons which then determines the truth of literal $H$. Thus we can use the network to determine the truth of $H$, call this $N(x)$. Furthermore, if we have some rule $R$ relating variables $H$ and $X_i$, we can use the state of the input $x$ to determine the value of the variables $X_i$, and then use $R$ to determine the value of $H$, call this $R(x)$. Given a set of input configurations to test $I$ (not necessarily from the test set of the network) we can measure the discrepancy between the output of the rules and the network as
\begin{equation}
E(R):=\frac{1}{|I|}\sum\limits_{x\in I} |R(x)-N(x)|
\end{equation}
In other words, we measure the average error of the rules when trying to predict the output of the network over a test set.

Comprehensibility is more difficult to define as there is a degree of subjectivity. The approach we take is to look at the \textit{complexity} of a rule. Here, we think of complexity analogously to the Kolmogorov complexity which is determined by a minimal description. Thus we determine the complexity of a rule by the length of its body when expressed by a (minimal) rule in disjunctive normal form (DNF). For an \textit{M-of-N} rule, the complexity is simply $M{{N}\choose{M}}$, where $\choose$ denotes the binomial coefficient. For our experiments we measure complexity in relative terms by normalizing w.r.t. a maximum complexity. Given $N$ possible input variables, the maximum complexity is $\lceil{\frac{N+1}{2}}\rceil {{N}\choose {\lceil{\frac{N+1}{2}}\rceil}}$, where $\lceil \rceil$ denotes the ceiling function (rounding to the next highest integer). Finally in order to control for growth we take the logarithm giving the following normalized complexity measure. 
\begin{equation}
C(R) :=\frac{\log(M{{N}\choose{M}})}{\log(\lceil{\frac{N+1}{2}}\rceil {{N}\choose {\lceil{\frac{N+1}{2}}\rceil}})}
\end{equation}
As an example, suppose we have a simple perceptron with two binary visible units with weights $w_{1,1}=1$ and $w_{2,1}=-0.5$ and whose output has a bias of $1$. Then consider the rule $h=1 \leftarrow 1$-of-$\{x_1=1,\neg (x_2=1)\}$. Over the entire input space we see that $R(x) \neq N(x)$ only when $x_1=0$ and $x_2=1$ giving us an error of $0.25$. Furthermore, a $1-of-2$ rule is the most complex rule possible for $2$ variables as it has the longest DNF of any \textit{M-of-N} rule giving us a complexity of $1$.

Using Eqs. $(1)$ and $(2)$ we define a loss function for a rule $R$ as a weighted sum in which a parameter $\beta\in \mathbb{R}^+$ determines the trade-off between soundness and complexity.
\begin{equation}
L(R):=E(R)+ \beta C(R)
\end{equation}
By using a brute force search with various values of $\beta$ we are able to explicitly determine the relationship between the allowed complexity of a rule and its maximum accuracy. For $\beta=0$ the rule with the minimum loss will simply be the rule with minimum error regardless of complexity, and for $\beta$ large enough the rule with the minimum loss will be a rule with $0$ complexity, either a $1-of-1$ rule or one of the trivial rules which either always predicts true or always predicts false (these can be represented as  \textit{M-of-N} rules by $0-of-N$ and $N+1-of-N$ respectively). 

\subsection{Layerwise \textit{M-of-N} Rule Extraction Algorithm}

Given a neuron $h_j$ with $n$ input neurons $x_i$, we generate splits for each neuron using the technique just described to obtain a set of literals $H_j$ and $X_i$. Then, we negate the literals corresponding to neurons which have a negative weight to $h_j$. Using these we search through $\mathcal{O}(n^2)$ \textit{M-of-N} rules with variables $X_i$ in the body and $H_j$ in the head, which minimize $L(R)$. To do this, as a heuristic, we reorder the variables according to the magnitude of the weight connecting $x_i$ to $h_j$ (such that we have $|w_{1,j}|\geq |w_{2,j}| \geq ... \geq |w_{n,j}|$). Then we consider the rule $M-of-\{X_1,...,X_N\}$ for each $1\leq N \leq n$ and each $0 \leq M \leq N+1$. The search procedure only relies on the ordering of the variables $X_i$. 
By ordering the literals according to the magnitude of their weights we reduce an exponential search space to a polynomial one. In the ideal case the set of possible input values to a hidden neuron is $X^n$ (where $X$ is the set of values that each input neuron can possibly take); it can be easily proved that the weight-ordering will find an optimal solution. In practice however, certain inputs may be highly correlated. When this is the case there is no guarantee that the weight-ordering will find the optimal \textit{M-of-N} rule. Thus in the general case the search procedure is heuristic. This heuristic allows us to run our search in parallel. We do this by using Spark in IBM Watson studio.

To illustrate the entire process, let us examine rule extraction from the first hidden layer in the CNN trained on the fashion MNIST data set. First we randomly select a set of examples and use them to compute the activations of each neuron in the CNN as well as the predicted labels of the network. With padding there are $28\times28=784$ neurons per feature in the first hidden layer, each corresponding to a different $5\times 5$ patch of the input image. We then find the optimal splitting value of each neuron by computing the information gain of each splitting choice with respect to the network's predicted labels. We find that the neuron with the maximum information gain is neuron $96$ which has an information gain of $0.015$ when split on the value $0.0004$. This neuron corresponds to the image patch centered at $(3,12)$. With this split we define the variable $H$ as $H:= 1$ iff $h_{96} \geq 0.0004$. 

Using this variable we define the input splits by choosing the values which result in the maximum information gain with respect to $H$. We then search through the \textit{M-of-N} rules whose bodies consist of the input variables defined by the splits to determine an optimal \textit{M-of-N} rule explaining $H$ for various error-complexity tradeoffs. As we increase the complexity, three different rules are extracted which can be visualized in Figure 1. As can be seen, many of the weights are filtered out by the rules. The most complex rule is a $5$-of-$13$ rule which has a $0.025 \%$ error. A mild complexity penalty changes the optimal rule to the much simpler $3$-of-$4$ rule, but raises the error to $0.043\%$. And a heavy complexity penalty produces a $1$-of-$1$ rule which has the significantly higher error of $0.13\%$. 
\begin{figure}[t!]
  \centering
  \includegraphics[scale=15]{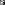}
  \includegraphics[scale=15]{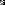}
  \includegraphics[scale=15]{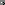}
  \includegraphics[scale=15]{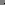}
  \caption{The leftmost image represents the weights of neuron $96$. The next three images are obtained from rules of decreasing complexity extracted from the CNN explaining that neuron. If a literal is true (resp. false) it is shown in white (resp. black). Grey indicates that the input feature is not present in the \textit{M-of-N} rule. Notice how a rule can be seen as a discretization of the network into a three-valued logic, similar to what is proposed by binarized networks \cite{DBLP:journals/corr/CourbariauxB16} but without constraining the network training a priori.}
  \vspace{-0.5cm}
\end{figure}

\section{Experimental Results}

\subsection{Small Fully Connected Networks} In order to compare our search procedure with CORELS as an optimal baseline \cite{CORELS}, we evaluate both methods on a series of small fully connected networks. The first is a deep neural network with $2$ fully connected layers of $16$ and $8$ hidden neurons, respectively, with a rectified linear (ReLu) activation function, on the car evaluation dataset \cite{car}. The second is a single layer network with $100$ hidden neurons with ReLU activations trained on the E. Coli dataset \cite{ECOLI}. The final network is a single layer $25$ hidden unit network trained on the DNA promoter dataset \cite{DNA}. Because the DNA promoter dataset is quite small, we produce $10,000$ synthetic examples to evaluate our rule extraction methods on the final network. We simply use the entire dataset for the other two networks.\\\\
CORELS produces optimal rules for a given set of parameters (maximum cardinality, minimum support and a regularization parameter) also seeking to penalize complexity. Maximum cardinality refers to the maximum number of literals in the body of a rule, the minimum support refers to the minimum number of training examples an antecedent must capture to be considered in the search, finally the regularization parameter is a scalar penalty on the complexity, equivalent to the parameter $\beta$ used in our \textit{M-of-N} search. 

Because our extraction algorithm uses an ordering on the literals, each rule can be evaluated independently so that the search procedure can run in parallel. This greatly speeds up the search compared to CORELS, which requires a sequential search. This faster search will allow us to apply the extraction algorithm to larger networks and to use more test examples. However, since we only search over \textit{M-of-N} rules we are not guaranteed to find an optimal solution. For this reason we compare our layerwise results with CORELS to see how far from optimal our search method is. Since CORELS has multiple parameters to penalize complexity we run CORELS multiple times with different parameters to generate a set of rules with higher complexity and one with lower complexity and then compare these rules to rules of similar complexity found by our parallel search.

In Table 1 we can see that rules found via our \textit{M-of-N} search are only marginally worse than a set of optimal rules with similar complexity found by CORELS and that CORELS can become quite slow when using too broad a search on a dataset with many inputs. When applied to the DNA promoter network CORELS runs out of memory and we were unable to produce a result showing that even for this relatively small network CORELS is too computationally demanding. Notice also that in this example the second hidden layer is much more explainable than the first, c.f. the large difference in accuracy between layers. 
\begin{table}[h]
\caption{Comparison of rules extracted from different layers of networks trained on various datasets using (sequential) CORELS with different values for cardinality/support/regularization and our Parallel \textit{M-of-N} extraction using different values for $\beta$. At a similar level of complexity (Comp), rules extracted by CORELS are only marginally more accurate (c.f. Acc) than \textit{M-of-N} rules, despite CORELS searching over a much larger sequential rule space; refer to computation time (Time). Below, n/a is used when CORELS exits without terminating.}
\centering
\begin{tabular}{l l l l l}
\hline
\textbf{Method} & \textbf{Comp} & \textbf{Acc} & \textbf{Network} & \textbf{Time}  \\
\hline
CORELS(1/0.01/0.01) & n/a & n/a & DNA promoter Layer 1 & n/a \\
Parallel \textit{M-of-N} \small{$\beta=0.3$} & 0.239 & 89\% & DNA promoter Layer 1 & 700s\\
\hline
CORELS \small{(1/0.01/0.01)} & 0.124 & 93.4\% & Cars Layer 1  & $1s$   \\ 
CORELS \small{(2/0.05/0.05)} & 0.04 & 87.3\% & Cars Layer 1 & $1800s$    \\ 
Parallel \textit{M-of-N} \small{$\beta =0.2$} & 0.131  & 90.3\% & Cars Layer 1 & $1s$ \\ 
Parallel \textit{M-of-N} \small{$\beta =1$} & 0.031 & 85.4\% & Cars Layer 1 &$1s$   \\ 
\hline
CORELS \small{(1/0.01/0.01)} & 0.053 & 99.05\%  & Cars Layer 2 & $1s$  \\ 
CORELS \small{(3/0.02/0.02)} & 0.079 & 99.42\% & Cars Layer 2 & $1s$\\
Parallel \textit{M-of-N} \small{$\beta =0.3$} & 0.057  & 98.4\% & Cars Layer 2 & $1s$ \\ 
Parallel \textit{M-of-N} \small{$\beta =0.1$} & 0.069 & 98.6\% & Cars Layer 2 & $1s$\\
\hline
CORELS \small{(1/0.01/0.01)} & 0.165 & 91.6\% & E.COLI Layer 1  & $1s$   \\ 
CORELS \small{(2/0.005/0.001)} & 0.287 & 92.6 \% & E.COLI Layer 1 & $10s$   \\
Parallel \textit{M-of-N} \small{$\beta =0.2$} & 0.132  & 89.4\% & E.COLI Layer 1 & $1s$ \\ 
Parallel \textit{M-of-N} \small{$\beta =0.1$} & 0.189  & 90.2\% & E.COLI Layer 1 & $1s$\\
\hline
\end{tabular}

\end{table}
Finally, the rate of accuracy decrease vs. complexity of Parallel \textit{M-of-N} seems to be lower than that of CORELS; this deserves further investigation. In summary, the above results show that a parallel \textit{M-of-N} search can provide a good approximation of the complexity/error trade-off for the rules describing the network. Next, we apply Parallel \textit{M-of-N} to much larger networks for which sequential or exhaustive methods become intractable.
\begin{figure}[t!]
\subfloat[CNN-Relu]{   \includegraphics[scale=0.3]{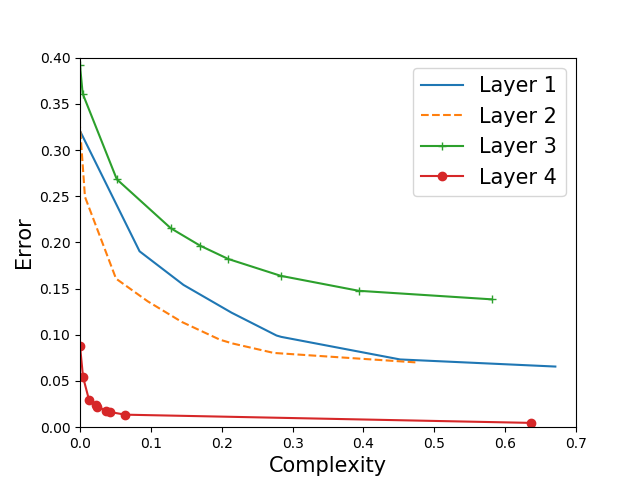}}
\subfloat[CNN-Tanh]{   \includegraphics[scale=0.3]{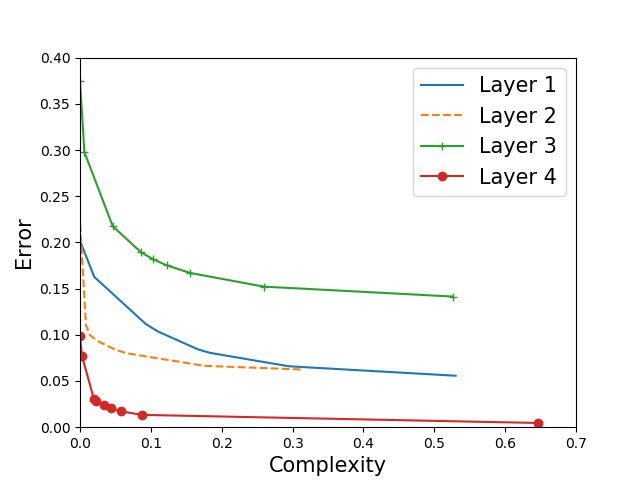}}
\subfloat[CNN-AE]{   \includegraphics[scale=0.3]{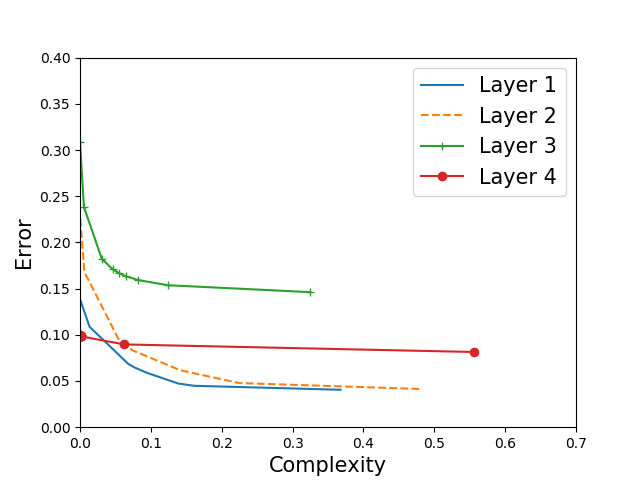}}
\caption{The Complexity/Error relationship for rules extracted from each layer of three different deep networks trained on MNIST. From left to right a CNN with relu activations trained end-to-end, a CNN with tanh activations trained end-to-end, a CNN with Relu activations trained as an autoencoder.}
\vspace{-0.5cm}
\end{figure} 
\subsection{Deep Convolutional Networks} In order to evaluate the capability of compact \textit{M-of-N} rules at explaining hidden features, we now apply the extraction algorithm to the hidden layers of three different networks trained on MNIST and compare results. Since applying extraction hierarchically can cause an accumulation of errors from previous layers, we use the network to compute the values of the inputs to the hidden layer that we wish to extract rules from. Hence, the errors from the extracted rules correspond to rule extraction at that layer. This allows us to examine the relative explainability at each layer. In practice, one could extract a hierarchical set of rules by choosing a single splitting value for each neuron. 

Our three networks are identical save for the activation function and training procedure. The network architecture consists of two convolutional layers with $16$ and $8$ filters respectively, each with a $3\times3$ convolutional window and using max pooling. This is followed by a $128$-unit densely connected layer with linear activation followed by a softmax layer. The first network uses ReLu units in the first two layers and is trained end-to-end. The second network is trained identically to the first but uses the hyperbolic tangent (Tanh) activation function in the first two layers. The third network uses an autoencoder to train the first three layers unsupervised before training the final softmax layer separately. We evaluate the rules using $5000$ examples from the test set

Comparing the network using ReLu to the one using Tanh shows that in both cases the minimum error for each layer remains approximately the same. However, the explainability in the Tanh network is greatly increased in the first three layers, rules extracted from the Tanh network can be made much less complex without significantly increasing the error. This applies not only to the first two layers but also to layer 3 which uses a linear activation in both cases. In both cases the third layer is much less explainable than the first two and the only layer which we are truly able to produce an acceptably accurate and comprehensible explanation is the final one in which we see rules with an average complexity of $0.087$ achieving an average error of $0.013\%$.

In the third layer we believe that the higher minimum error is mainly the result of the number of input units. In these layers there appear to be a lot of input units which are not
relevant enough alone to be included in an \textit{M-of-N} rule, but collectively they add enough noise to have a significant effect on the output. Because our search procedure is heuristic, it's possible that a more thorough search could produce rules which are simpler and 
more accurate but our results at least tentatively back up the idea that the distributed nature of neural networks makes rule extraction from the hidden layers impractical if not infeasible. We hypothesize that the difference in complexity between rules extracted from the Tanh network and the Relu network is due to the saturating effect of the tanh function. A hidden neuron in the tanh network may have fewer `marginally relevant' features than in the Relu network. This would explain the steep decline in accuracy found in the Tanh network and the more gradual decline found in the Relu network.

The autoencoder has hidden features which are in general more explainable than either of the two previous networks. Compared to the ReLu network, the error of the extracted rules in the second layer is lower at every level of complexity. Compared to the Tanh network, the autoencoder has more accurate rules at medium levels of complexity ($6.1\%$ error at $0.144$ complexity vs. $6.6\%$ error at $0.18$ complexity). However, as complexity is reduced the extracted rules in the Tanh network remain accurate for longer ($9.6\%$ error at $0.053$ complexity vs. $8.4\%$ at $0.048$ complexity). Interestingly, in the autoencoder the second layer is slightly less explainable than the first. The third layer is more explainable than it is in the other two networks with significant increases in error only being seen with rules of average complexity less than $0.08$. In the softmax layer trained on top of the autoencoder we see that one cannot extract accurate rules of any complexity.
This points to something fundamentally different from the previous two networks in the way that softmax uses the representations from the final layer to predict the output. This is the subject of further investigation. 

Our results indicate that, at least when it comes to extracting \textit{M-of-N} rules with an assumption of weight-ordering, there are hard limitations to representing hidden units that cannot be overcome with any level of complexity. These limitations seem to be the result of the internal representations determined by the training procedure. Whether these limitations can be overcome by refining rule extraction methods or whether they are a fundamental part of the network is to be determined. However, we also find that the final layer of a CNN may be a promising target for rule extraction. We verify this by training $2$ more 4-layer CNNs on the Olivetti faces and fashion MNSIT dataset. The network trained on the Olivetti faces dataset consists of two convolutional layers with $20$ and $10$ filters respectively each with a $3 \times 3$ window and followed by $2 \times 2$ max pooling. Then a $256$ unit fully connected hidden layer with a linear activation followed by the softmax layer. The fashion MNIST network is larger. It has two convolutional layers with $32$ and $64$ filters with a $5 \times 5$ window followed by $2 \times 2$ max pooling. Then a $1024$ unit fully connected layer followed by the softmax. Olivetti faces is evaluated using the entire dataset and fashion MNIST is evaluated with $1000$ samples.

In Table 2 we can see that the Olivetti Faces dataset had the most accurate and interpretable rules of all, this is probably at least partially due to the smaller size of the dataset. In all cases one can see a large drop in the complexity with only a penalty of $\beta=0.1$ resulting in a less than $1\%$ decrease in accuracy. This suggests that in the softmax layer, relatively few of the input neurons are being used to determine the output. This shows that rule extraction, and in particular \textit{M-of-N} rule extraction can be an effective component in a multi-pronged approach to explainability. By extracting \textit{M-of-N} rules from the final layer and using importance methods to explain the relevant hidden units, one should be able to reason about a network's structure in ways that cannot be achieved with a strictly model-agnostic approach. Such a hybrid approach is expected to create explanations which can be accurate and yet less complex. 

\begin{table}[t!]

\centering
\caption{Comparison of the complexity (Comp), and accuracy (Acc) of rules extracted from the final layer of three CNNs trained on different datasets. Repeated for complexity penalties of $\beta=0$ and $\beta=0.1$}
\begin{tabular}{l l l l l}
\hline
\textbf{Dataset} & \textbf{Comp. ($\beta=0$)} & \textbf{Acc.($\beta=0$)} & \textbf{Comp. ($\beta=0.1$)} & \textbf{Acc.($\beta=0.1$)}   \\
\hline
Olivetti Faces  & 0.03 & 100\% & 0.024 & 99.9\%    \\ 
MNIST  &0.7 & 99.6\% & 0.06 & 98.7\%    \\ 
Fashion MNIST & 0.28  & 99.3\% & 0.06 & 98.8\%  \\ 
\hline
\end{tabular}
\vspace{-0.5cm}
\end{table}

\vspace{-0.3cm}
\section{Conclusion and Future Work}
\vspace{-0.3cm}
 The black box problem has been an issue for neural networks since their creation. As neural networks become more integrated into society, explainability has attracted considerably more attention. The success of knowledge extraction in this endeavor has overall been mixed with most large networks today remaining difficult to interpret and explain. Traditionally, rule extraction has been a commonly used paradigm and it has been applied to various tasks. Critics, however, point out that the distributed nature of neural networks makes the specific method of decompositional rule extraction unfeasible as individual latent features are unlikely to represent anything of significance. We test this claim by applying a novel search method of \textit{M-of-N} rule extraction to generate explanations of varying complexity for hidden neurons in a deep network. We find that the complexity of neural representations does provide a barrier to comprehensible rule extraction from deep networks. However we also find that within the softmax layer rule extraction can be both highly accurate and simple to understand. This shows that rule extraction, including \textit{M-of-N} rule extraction can be a useful tool to help explain parts of a deep network. As future work, softmax layer rule extraction will be combined with local explainability techniques. Additionally, our preliminary experiments suggest that replacing the output layer of a network with \textit{M-of-N} rules may be more robust to certain adversarial attacks. Out of $1000$ adversarial examples generated using FGSM \cite{FGSM} for the CNN trained on MNIST, $376$ were classified correctly by the \textit{M-of-N} rules with maximum complexity by contrast with none classified correctly by the CNN. This is to be investigated next in comparison with various other defense methods.   

\bibliographystyle{plain}
\bibliography{LayerwiseExtraction}

\end{document}